\documentclass{article}
\usepackage{spconf,amsmath,graphicx}

\title{% SAM Wears Spatio-Temporal, Bidirectional Audio-Visual Attention 
EXTENDING SEGMENT ANYTHING MODEL INTO AUDITORY AND TEMPORAL DIMENSIONS FOR AUDIO-VISUAL SEGMENTATION
}
\name{Juhyeong Seon, Woobin Im, Sebin Lee, Jumin Lee, and Sung-Eui Yoon*
\thanks{*Corresponding Author: Sung-Eui Yoon.
This work was supported by NRF and IITP grant funded by the Korea government(MSIT) (RS-2023-00208506(2024) and RS-2023-00237965(2024)).
}
}
\address{Author Affiliation(s)}
\address{School of Computing, Korea Advanced Institute of Science and Technology (KAIST), Daejeon, Korea}
\usepackage[dvipsnames]{xcolor}

\usepackage{caption}
\usepackage{afterpage}
\usepackage{algorithm}
\usepackage{algpseudocode}
\usepackage{graphicx}
\usepackage{amsmath}
\usepackage{amssymb}
\usepackage{rotating}
\usepackage{array}

\usepackage{booktabs}
\usepackage{tikz}
\usetikzlibrary{positioning}
\usepackage{multirow}
\usepackage{makecell}
\usepackage{colortbl}
\usepackage{listings}
\usepackage{pythonhighlight}
\usepackage{tabularx} % in the preamble
\usepackage{subcaption}
\begin{document}
\ninept
\maketitle
\begin{abstract}
Audio-visual segmentation (AVS) aims to segment sound sources in the video sequence, requiring a pixel-level understanding of audio-visual correspondence.
As the Segment Anything Model (SAM) has strongly impacted extensive fields of dense prediction problems, prior works have investigated the introduction of SAM into AVS with audio as a new modality of the prompt.
Nevertheless, constrained by SAM's single-frame segmentation scheme, the temporal context across multiple frames of audio-visual data remains insufficiently utilized.
To this end, we study the extension of SAM's capabilities to the sequence of audio-visual scenes by analyzing contextual cross-modal relationships across the frames.
To achieve this, we propose a Spatio-Temporal, Bidirectional Audio-Visual Attention (ST-BAVA) module integrated into the middle of SAM's image encoder and mask decoder.
It adaptively updates the audio-visual features to convey the spatio-temporal correspondence between the video frames and audio streams.
Extensive experiments demonstrate that our proposed model outperforms the state-of-the-art methods on AVS benchmarks, especially with an 8.3\% mIoU gain on a challenging multi-sources subset.
\end{abstract}
\section{INTRODUCTION}
\label{sec:intro}
\begin{figure}[!ht]
  \centering
  \begin{subfigure}[b]{0.225\textwidth}
   \includegraphics[trim={25cm 34cm 45cm 12cm}, clip=true, width=1.0\linewidth]{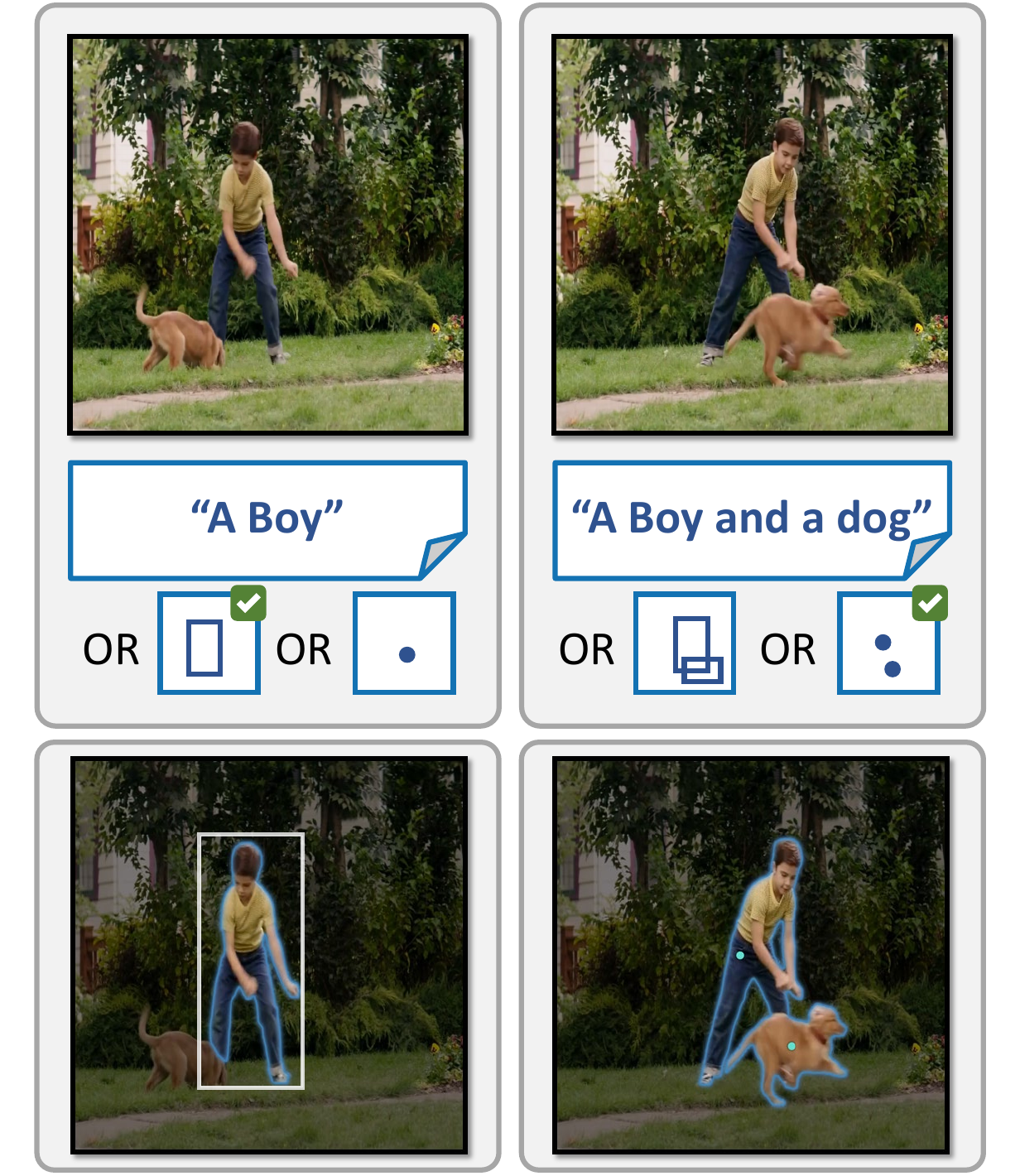}
    \caption{SAM}
   \label{fig:intro_sam}
  \end{subfigure}
  \hfill
  \begin{subfigure}[b]{0.225\textwidth}
   \includegraphics[trim={46cm 34cm 24cm 12cm}, clip=true, width=1.0\linewidth]{figures/figure_intro_v7.3.pdf}
    \caption{Prior SAM-based AVS~\cite{liu2023annotation, wang2023prompting}}
   \label{fig:intro_sama}
  \end{subfigure}
  \hfill
  \begin{subfigure}[b]{0.465\textwidth}
   \includegraphics[trim={27.6cm 6cm 27.5cm 37cm}, clip=true, width=1.0\linewidth]{figures/figure_intro_v7.3.pdf}
   % \vspace{-1mm}
    \caption{SAM with ST-BAVA (Ours)}
   \vspace{-1mm}
   \label{fig:intro_stbava}
  \end{subfigure}
   \label{fig:intro}
   \caption{
   Segmentation results of different models on a film \textit{A Dog's Purpose} (2017).
   (a) Segment Anything Model (SAM) segments target objects in the image with their regions guided by user prompts.
   (b) Prior works~\cite{liu2023annotation, wang2023prompting} have adapted SAM to segment objects that sound with a corresponding audio prompt per frame.
   (c) We propose a spatio-temporal, bidirectional audio-visual attention (ST-BAVA), enabling SAM to fully leverage the relationships between the subsequent video frames and audio streams in a bidirectional way. In Fig.~\ref{fig:intro_stbava}, our model successfully segments the human and the dog on the frames where they make sounds.}
   \label{fig:introduction}
    \vspace{-3mm}
\end{figure}

Segment Anything Model (SAM)~\cite{kirillov2023segment} is a foundation model in image segmentation with points, boxes, and text prompts (Fig.~\ref{fig:intro_sam}).
Tremendous works have shown SAM's outstanding performance in various dense prediction problems~\cite{cen2023segment, shen2023anything, wu2023medical, mazurowski2023segment, wang2023detect, zhang2023sam, cheng2023segment, li2023refsam, rajivc2023segment} by adapting it to specific domains.
For instance, Cen \textit{et al.}~\cite{cen2023segment} extended the segmentation ability of SAM to 3D scenes through NeRFs.
Wu \textit{et al.}~\cite{wu2023medical} adapted SAM into the medical domain, showing generalized segmentation performance on CT, MRI, ultrasound, fundus, and dermoscopic images.

% ================================================
In light of the success of SAM in various tasks, pioneering works~\cite{mo2023av, liu2023annotation, wang2023prompting} have attempted to introduce a prompt in the novel modality, audio, into the SAM (Fig.~\ref{fig:intro_sama}).
They aim to segment the object that makes the sound in the audible video, defined as Audio-Visual Segmentation (AVS) problem~\cite{zhou2022audio}.
Mo \textit{et al.}~\cite{mo2023av} explored a spatial fusion of the audio-visual features for audio prompting of SAM.
Liu \textit{et al.}~\cite{liu2023annotation} and Wang \textit{et al.}~\cite{wang2023prompting} presented prompt tuning techniques~\cite{houlsby2019parameter} by inserting the light-weight adapters into the image encoder and decoder of SAM respectively, achieving high performance on the AVS.

Nevertheless, since they divide the video into individual frames in a single-frame manner constrained by SAM, the contextual information provided by the audio-visual scene across subsequent frames has been neglected.
Furthermore, while the effectiveness of bidirectional modeling with the interaction between the image frame and the corresponding prompt has been demonstrated in the SAM-based referring video object segmentation (RVOS)~\cite{li2023refsam} and tracking~\cite{rajivc2023segment}, it has not been sufficiently explored for the audible video data.
As the AVS requires comprehensive pixel-level correspondence across two different modalities, leveraging the complementary relationship between audio-visual cues in a bidirectional way becomes more essential to this task.

\begin{table}[t]
  \resizebox{\linewidth}{!}{%
    \centering
    \begin{tabular}{p{1.9cm}l|cccc}
    \hline
    \multirow{2}{*}{Backbone} & \multirow{2}{*}{Methods} & \multicolumn{2}{c}{S4} & \multicolumn{2}{c}{MS3}  \\ \cline{3-6} 
      & & mIoU      & F-score    & mIoU      & F-score \\ \hline
    \multirow{7}{*}{PVTv2} & TPAVI~\cite{zhou2022audio} & 78.74      & 0.879      & 54.00      & 0.645   \\
    & AVSC~\cite{liu2023audiovisual}     & 81.29     & 0.886      & 59.50      & 0.657   \\
    & CATR~\cite{li2023catr}     & 81.40     & 0.896      & 59.00      & 0.700   \\
    & AQFormer~\cite{huang2023discovering}     & 81.60     & 0.894      & 61.10      & 0.721   \\
    & AVSegFormer~\cite{gao2023avsegformer}     & 82.06     & 0.899      & 58.36      & 0.693   \\
    & ECMVAE~\cite{mao2023multimodal}     & 81.74     & 0.901      & 57.84      & 0.708   \\
    & LDM~\cite{mao2023contrastive}      & 81.38     & 0.902      & 58.18     & 0.709   \\ \hline
    \multirow{3}{*}{SAM} & GAVS~\cite{wang2023prompting} & 80.06     & 0.902      & 63.70     & 0.774   \\
    & SAMA-AVS~\cite{liu2023annotation} & 81.53     & 0.886      & 63.14     & 0.691   \\
    & \cellcolor[HTML]{EFEFEF}ST-BAVA (Ours) & \cellcolor[HTML]{EFEFEF}\textbf{82.46} & \cellcolor[HTML]{EFEFEF}\textbf{0.906} & \cellcolor[HTML]{EFEFEF}\textbf{69.01} & \cellcolor[HTML]{EFEFEF}\textbf{0.776} \\ \hline
    \end{tabular}
  }
  \caption{Quantitative comparison with other AVS methods on the Single-source (S4) and Multi-source (MS3) subset of AVSBench dataset~\cite{zhou2022audio} regarding mIoU and F-score.}
  \label{tab:main_result_v2}
  \vspace{-3mm}
\end{table}
% 다시 쓰기?
To this end, we study the extension of SAM's segmentation capabilities to the subsequent frames with corresponding audio, proposing a Spatio-Temporal, Bidirectional Audio-Visual Attention module (ST-BAVA) (Fig.~\ref{fig:intro_stbava}).
It aims to convey the spatio-temporal relationships between input images and audio prompts to SAM through bidirectional adjustments of audio-visual features in the middle of the pipeline.
Our proposed method exhibits two advantages compared to the previous approaches~\cite{liu2023annotation, mo2023av, wang2023prompting}.
First, we enable SAM to leverage the contextual information presented in the audio-visual scene of multiple frames.
Second, we mutually align the image and audio features based on their spatial and temporal correspondences across the video.
Through the bidirectional aggregation of the spatio-temporal relationship across the sequence of the audio-visual frames, our model effectively identifies the distinct visual and auditory cues from the objects that could emerge or disappear.
Extensive experimental results (Sec.~\ref{sec:experiments}, \ref{sec:results}) demonstrate that the proposed method achieves high performance in localizing the sound source with only a few trainable parameters ($<$4\% of SAM), thanks to the audio-visual relationship exploited by ST-BAVA.
In particular, it achieves improvements of 8.3\% in mIoU on the AVS benchmark's multi-sources subset, as shown in Table~\ref{tab:main_result_v2}.

We summarize our contributions as follows: \begin{itemize}
    \item We extend SAM into the auditory and temporal dimensions to segment the sound sources on the subsequent video frames using corresponding audio as a prompt.
    \item We propose a Spatio-Temporal, Bidirectional Audio-Visual Attention module (ST-BAVA), enabling SAM to exploit the spatio-temporal correspondences between subsequent image frames and audio streams.
    \item Through experiments, we demonstrate that the proposed method outperforms the state-of-the-art methods on the AVS. 
    Furthermore, we showcase the effectiveness of the main components in our approach through extensive ablation studies.
\end{itemize}

\section{RELATED WORK}
\label{sec:related_work}
\subsection{Audio-Visual Segmentation}
In audio-visual learning, the relationship between audio and visual data has been explored to understand the scene in multimodal.
Researchers pioneered the audio-visual correspondence (AVC)~\cite{aytar2016soundnet, arandjelovic2017look} with a binary classification to predict whether the image and audio data correspond.
The relationship exploited in AVC has been extended into the spatial dimension, evolving into the task named sound source localization (SSL) that aims to localize the region of sound sources in the frame~\cite{zhao2018sound, chen2021localizing, mo2023audio}.
With a remarkable achievement of SSL, Zhou \textit{et al.} ~\cite{zhou2022audio} introduced an advanced segmentation challenge called Audio-Visual Segmentation (AVS).
Diverse approaches have focused on learning effective multimodal representations to comprehend the pixel-level audio-visual correspondences in the video scene.
Mao \textit{et al.}~\cite{mao2023multimodal, mao2023contrastive} introduced the multimodal VAE and conditional latent diffusion model to learn the advanced audio-visual representation.
Huang \textit{et al.}~\cite{huang2023discovering} and Gao \textit{et al.}~\cite{gao2023avsegformer} utilized the transformer-based architecture with the interaction between audio-conditioned object queries and visual features.
Liu \textit{et al.}~\cite{liu2023audiovisual} proposed the two-stage framework: segmenting all potential objects from the visual data and verifying sounding objects using an audio-visual semantic matching.
In this work, we introduce a bidirectional cross-modal feature interaction module to extend the capabilities of the Segment Anything Model (SAM)~\cite{kirillov2023segment} to the AVS.

\begin{figure*}[t]
  \centering
  \begin{subfigure}[b]{0.55\textwidth}
   \includegraphics[trim={1.5cm 5.3cm 25cm 6.85cm}, clip=true, width=1.0\linewidth]{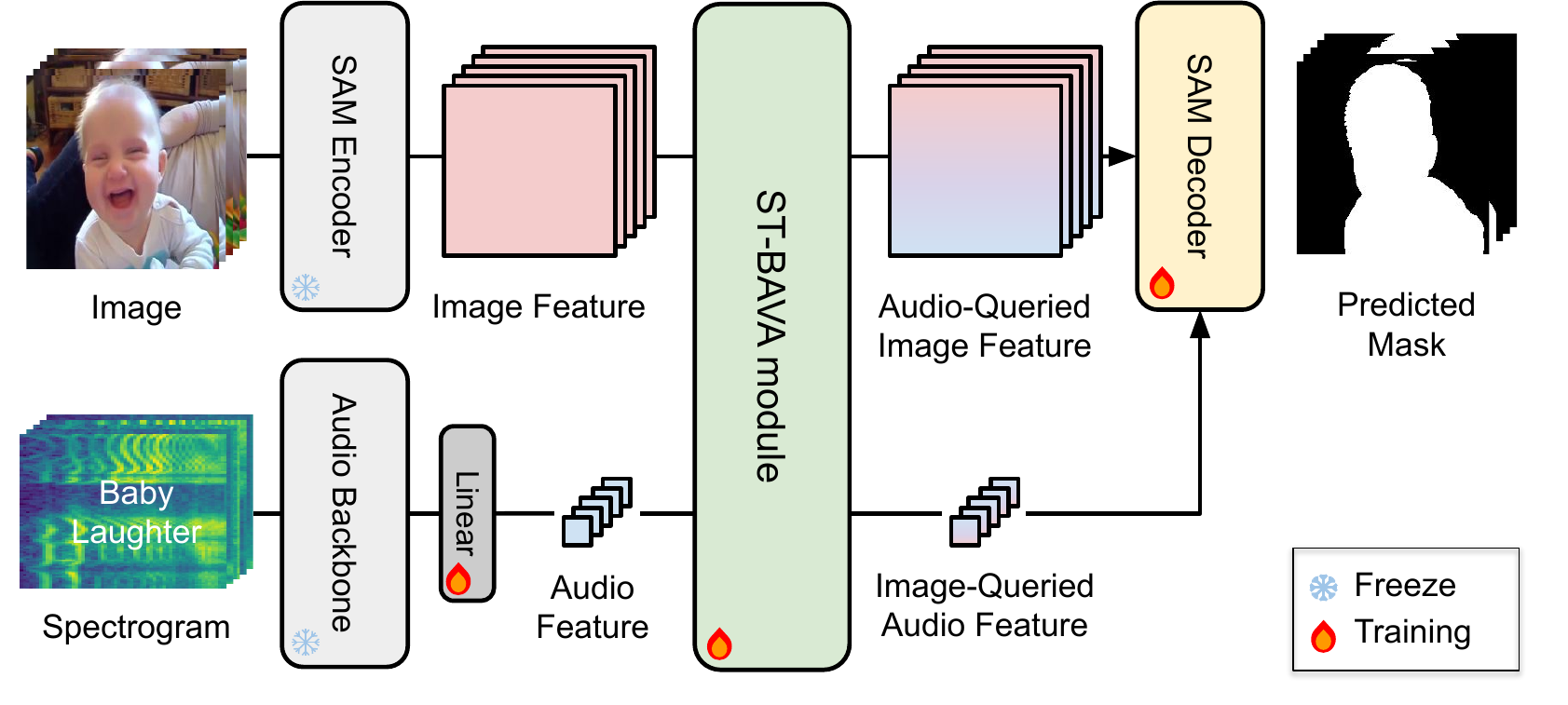}
   \caption{Pipeline of SAM with ST-BAVA}
   \label{fig:main_method_pipeline}
    \vspace{-1mm}    
  \end{subfigure}
   \hfill
  \begin{subfigure}[b]{0.43\textwidth}
   \includegraphics[trim={30.5cm 5cm 3cm 6.2cm}, clip=true, width=1.0\linewidth]{figures/figure_method_v5.2.pdf}
   \caption{ST-BAVA module architecture}
   \label{fig:main_method_ST_BAVA}
    \vspace{-1mm}    
  \end{subfigure}
   \caption{Overview of the proposed SAM with ST-BAVA.
   (a) 
   Our model takes a sequence of video frames and audio streams as input and predicts the mask of the sound sources for each video frame.
   (b)
   ST-BAVA module bidirectionally updates the image and audio features with spatial and temporal attention in sequence. M.H. stands for the multi-head. The initial audio feature from the audio backbone is used as a positional encoding for the audio feature.}
   \label{fig:main_method}
    \vspace{-3mm}    
\end{figure*}

\subsection{Segment Anything Model}
SAM is a foundation model for generality and broad applications in image segmentation problems with point, box, mask, and text prompts, pre-trained on 1B masks from 11M images~\cite{kirillov2023segment}.
Extensive works have studied the SAM's ability with various problems, including 3D vision tasks~\cite{cen2023segment, shen2023anything}, 
medical image segmentation~\cite{wu2023medical, mazurowski2023segment}, and shadow detection~\cite{wang2023detect, zhang2023sam}.
Within this context, the extension of SAM to handle the video data has been explored~\cite{rajivc2023segment, li2023refsam, cheng2023segment}.
For instance, Cheng \textit{et al.}~\cite{cheng2023segment} proposed a user-interactive video object tracking framework by supporting SAM's insufficient temporal and semantic understanding of the object.
They employed Grounding Dino~\cite{liu2023grounding}, a vision-language model, to interactively convert the user's description into the box prompt for SAM.
Beyond these approaches, our work introduces SAM's temporal extension with audio as a new input modality without relying on the existing prompts.

\section{METHODOLOGY}
\label{sec:method}
In this work, we introduce SAM to solve the audio-visual segmentation problem, where the model predicts the pixel-level segmentation mask of sounding objects on the video frames with the corresponding audio prompts (Sec.~\ref{sec:method_p.f.}).
After revisiting the image segmentation process of SAM as a preliminary (Sec~\ref{sec:sam_revisit}), we discuss how to extend the SAM into the auditory and temporal dimensions (Sec.~\ref{sec:sam_extend}) and introduce our proposed method ST-BAVA (Sec.~\ref{sec:method_sam_pipeline},~\ref{sec:method_st_bava}).
Finally, Adapter~\cite{liu2023annotation} to further aid effective audio-visual fusion of SAM in the image encoding is introduced (Sec.~\ref{sec:adapter}).

\subsection{Problem Formulation}
\label{sec:method_p.f.}
The video data for the audio-visual segmentation consists of a series of visual frames and audio spectrograms. 
For the video $n$ with a length of $T$ seconds, denoted as ($I^{n}, S^{n}$), the image frames are represented as 
$I^{n} = \left\{ i_{t}^{n} \right\} _{t=1}^{T}$
with an image $i_{t}^{n}\in \mathbb{R}^{3\times H_i\times W_i}$ at timestep $t$.
Each video frame is extracted at the end of each second.
The audio spectrograms are represented as 
$ S^{n} = \left\{ s_{t}^{n} \right\} _{t=1}^{T}$ with 
$ s_{t}^{n}\in \mathbb{R}^{H_{s}\times W_{s}}$. The spectrogram is processed via short-time Fourier transform of the 1-second audio clip.
The model outputs the segmentation map of the sound source as a binary mask $Y^{n} = \left\{ y_{t}^{n} \right\} _{t=1}^{T}$ with $y_{t}^{n}\in \left\{0, 1 \right\}^{H_i\times W_i}$. 
Each pixel in $y_t^n$ represents whether it is a sounding object. Multiple sound sources can be depicted within the single mask $y_{t}^{n}$ without considering semantic differences for each object.
The target segmentation mask is also provided as a binary mask $M^{n} = \left\{ m_{t}^{n} \right\} _{t=1}^{T}$ with $m_{t}^{n}\in \left\{0, 1 \right\}^{H_i\times W_i}$.
For simplicity, we omit the notation $n$ afterward.

\subsection{Revisiting SAM}
\label{sec:sam_revisit}
SAM is an architecture designed to solve the image segmentation problem.
It has three main components: an image encoder, a prompt encoder, and a mask decoder.
For the image encoder, a Vision Transformer (ViT)~\cite{dosovitskiy2020image} pre-trained with MAE~\cite{he2022masked} is used to extract the spatial image features.
A prompt encoder supports two subsets of prompts: sparse (points, boxes, and text) and dense prompts (masks).
% The points are represented as the summation of positional embeddings of the point's location and the learned embeddings.
% The boxes are similarly represented as points, with their top-left and bottom-right corner treated as two points.
% For the text, embeddings from the text encoder of CLIP~\cite{radford2021learning} are used. 
% Unlike these sparse embeddings, the mask is embedded using convolution operations.
These prompt embeddings are fed into the mask decoder, providing the regional information about the target objects to segment.
After the dense embedding is summed with the image embedding, the decoder operates the cross-attention between the image and the sparse prompt embeddings to update both.
With two iterations of these attentions operated, the dot product of upsampled image embedding and MLP-mapped sparse prompt embeddings outputs the final segmentation map.

\subsection{Extending SAM into Auditory and Temporal Dimensions}
\label{sec:sam_extend}
While SAM utilizes visual prompts to address various image segmentation problems, enabling SAM to support the audio prompts within multi-frame video requires an alternative approach.
In the pipeline of SAM, the guidance of prompts about where to segment is done in the lightweight mask decoder with cross attention between the image and prompt embeddings.
Considering this, a straightforward approach to processing the audio in SAM is to forward the audio embeddings as the prompt into the decoder, as introduced by Liu \textit{et al.}~\cite{liu2023annotation}.
However, using audio to represent the target objects in images is more complicated than using points and boxes, making it challenging for the lightweight decoder to comprehend the intricate audio-visual relationship~\cite{liu2023annotation}.
Moreover, as the decoder operates the cross attention per frame, the temporal dependencies among subsequent frames are insufficiently leveraged.
To solve these challenges, we propose the intermediate audio-visual interaction module ST-BAVA before the decoder.
This module aggregates the spatio-temporal relationship of the audio-visual features extracted from the subsequent video frames and audio streams.
These updated features by ST-BAVA are handed to the subsequent decoding process of the segmentation map, enabling the SAM's decoder to utilize the audio-visual correspondence in spatial and temporal dimensions.

\subsection{SAM with ST-BAVA}
\label{sec:method_sam_pipeline}
Fig.~\ref{fig:main_method_pipeline} shows the pipeline of SAM with the proposed ST-BAVA module. SAM's image encoder embeds input images $I$ to get the visual embedding $V\in \mathbb{R}^{T\times HW\times C}$, where $H, W$ represent the spatial size of the visual embeddings and $C$ represents a channel dimension.
The audio backbone encoder, followed by the learnable linear layer, encodes the audio spectrograms $S$ to the audio embedding $A\in \mathbb{R}^{T\times C}$, aligning with the channel dimension of visual embedding.
$V$ and $A$ are forwarded to the ST-BAVA, where the spatial and temporal attention operates sequentially.
Spatial attention computes the spatial correspondence between $V$ and $A$ for each time step, and temporal attention analyzes their audio-visual correlation across consecutive frames per pixel.
By the bidirectional operation of the spatial and temporal attention, ST-BAVA produces the audio-queried visual embedding $V_{\text{aq}}$ and visual-queried audio embedding $A_{\text{vq}}$.
Both are inserted into the SAM's mask decoder where $V_{\text{aq}}$ serves as the dense prompt and $A_{\text{vq}}$ as the sparse prompt, replacing the inherent prompts in SAM (Sec.~\ref{sec:sam_revisit}).
The following decoding steps are carried out in the same way as in image segmentation.
As the objective function for the AVS, we adopt the binary cross-entropy loss with the ground truth pixel mask $M$ for prediction $Y$, represented as $\mathcal{L} = \text{BCE}(Y, M)$.

\subsection{Spatio-Temporal, Bidirectional Audio-Visual Attention}
\label{sec:method_st_bava}
The design of ST-BAVA is motivated by the SAM decoder's bidirectional image-prompt feature fusion, extending it into spatio-temporal dimensions between the sequence of image and audio embeddings.
As processing audible video data requires high computational costs~\cite{wang2023prompting}, ST-BAVA uses the decomposed spatial and temporal cross-modal attention to reduce the memory requirements~\cite{bertasius2021space}.

Fig.~\ref{fig:main_method_ST_BAVA} elaborates ST-BAVA module architecture.
The temporal attention weight $\alpha_{\text{time}}$ exploits the contextual relationship between the audio-visual features per pixel, represented as follows:
\begin{align}
\label{eqn:temporal_attention_vtoa}
\alpha_{\text{time}} = \text{softmax}(\frac{VA^{\top}}{\sqrt{C}})\in \mathbb{R}^{HW \times T \times T}.
\end{align}
Note that the time axis $\mathbb{R}^T$ is used as the sequence dimension and the spatial axis $\mathbb{R}^{HW}$ as the batch.
To match the shape of audio and visual features, spatial average pooling and repeating the input and output of the attention are applied.
In spatial attention, the score map $\alpha_{\text{space}}$ calculates the spatial correspondence between $V_q$ and $A_{k}$ at each time step:
\vspace{-3mm}
\begin{align}
\label{eqn:spatial_attention_vtoa}
\alpha _{\text{space}} = \text{softmax}(\frac{VA^{\top}}{\sqrt{C}}) \in \mathbb{R}^{T \times HW \times{1}},
\end{align}
with the spatial axis $\mathbb{R}^{HW \times{1}}$ used as the sequence dimension.
This sequential operation of ST-BAVA first produces the image-queried audio embedding $A_{\text{vq}}$ and uses it to produce the audio-queried image embedding $V_{\text{aq}}$.
Repeated $L$ times, the ST-BAVA blocks aggregate the bidirectional audio-visual relationship across the spatio-temporal dimensions to support the subsequent decoding process.

\begin{figure}[t]
  \centering
   \includegraphics[trim={8.3cm 5.8cm 11.4cm 5.4cm}, clip=true, width=0.93\linewidth]{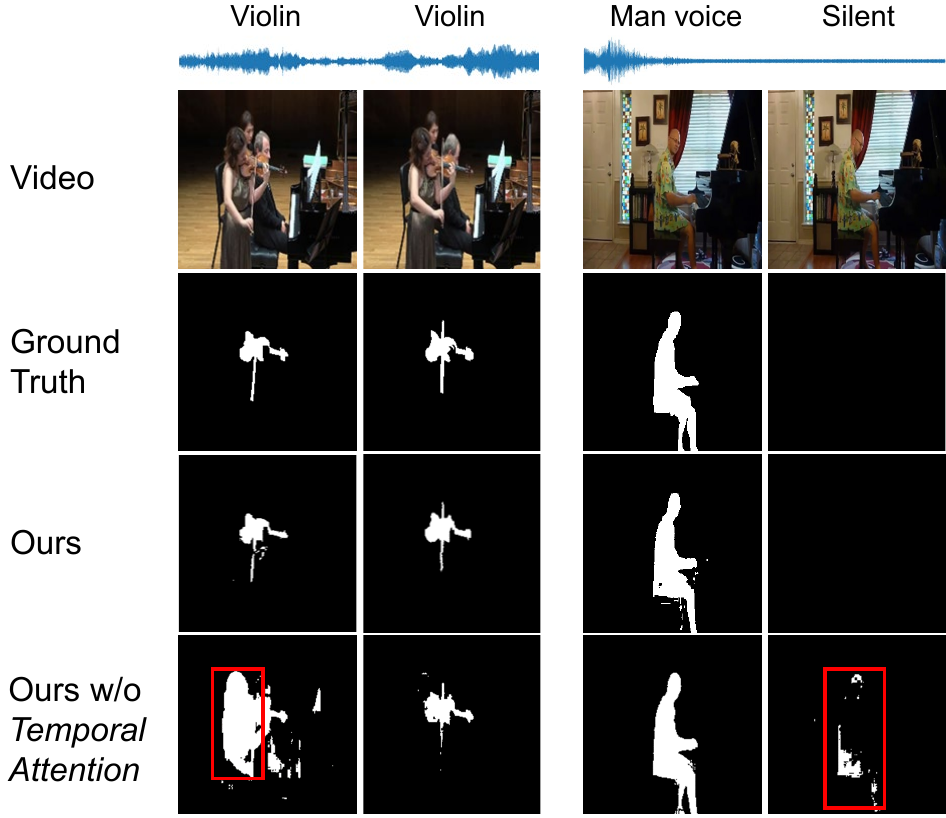}
   \caption{Effect of temporal attention in ST-BAVA on the audio-visual segmentation results.
   Our model leverages the temporal relationship across multiple frames, leading to accurate sound source predictions.
   Wrong prediction without temporal attention is marked in red boxes.
   }
   \label{fig:method_temp}
    \vspace{-3mm}    
\end{figure}

Fig.~\ref{fig:method_temp} shows the effect of temporal attention in the proposed ST-BAVA on leveraging contextual information across multiple frames. This leads to more accurate predictions of the sound source among visual candidates (violin in the left video), or judging the silent frame with no prediction (second frame of the right video) than the results without temporal attention. 

\subsection{Adapters} 
\label{sec:adapter}
Although our proposed ST-BAVA effectively adjusts the audio-visual features at the middle of the pipeline, it relies exclusively on the visual embeddings extracted from the SAM's pre-trained image encoder that operates independently of the audio data.
As Liu \textit{et al.}~\cite{liu2023annotation} have revealed, audio-visual feature fusion during the image encoding stage can further enhance AVS performance.
To achieve this, they have proposed Adapters that inject the audio feature into the image encoder.
In this work, we introduce the audio adapters to assist the ST-BAVA module's effective feature fusion with the audio-injected image features by Adapters.
Note that the Adapter does not utilize contextual or bidirectional relationships between multiple frames, as it injects audio information into the corresponding image.

For the detailed methods, in the $j$-th transformer image encoder layer stage, the $j$-th Adapter encodes the audio embedding $A$ from the audio backbone into $A^{prompt}_j\in \mathbb{R}^{T\times HW\times C}$ that has repeated spatial dimension. It is added to the output of the previous encoder layer $E_{j-1}$ to be the input of $j$-th layer $X^j$, represented as:
\begin{align}
X_j = E_{j-1}(X_{j-1}) + A^{prompt}_j.
\end{align}
The image embedding generated by the encoder with the Adapters proceeds through the ST-BAVA and mask decoder.

\section{EXPERIMENTAL SETUP}
\label{sec:experiments}
%------------------------------------------------------------------------
\subsection{Dataset}

We use the AVSBench dataset~\cite{zhou2022audio} designed for the audio-visual segmentation. 
It contains two video subsets: Single Sound Source Segmentation (S4) includes 4,932 videos, and Multi Sound Source Segmentation (MS3) contains 424 videos.
In S4, a single sound source consistently appears in each video, whereas multiple sources can appear or vanish as frames progress in MS3.
The videos cover 23 categories, including human voice, playing instruments, etc.
Each 5-second video is split into five image frames captured per second and five audio segments, each lasting 1 second.

\subsection{Evaluation Metrics} 
We use the mean Intersection over Union (mIoU) and F-score $\mathcal{M_{F}}$ for the evaluation metrics following~\cite{zhou2022audio}.
The mIoU computes the mean IoU between the predicted mask and the ground truth of the five frames. Note that mIoU is also known as Jaccard index $\mathcal{M_{J}}$ in related works~\cite{zhou2022audio, liu2023audiovisual, huang2023discovering}. The F-score $\mathcal{M_{F}}$ is calculated with the precision and recall, represented as $\mathcal{M_{F}} = \frac{(1+\beta^{2})\times precision\times recall}{\beta^{2} \times precision+recall}$. $\beta^{2}$ is set to 0.3 in our experiments.

\subsection{Implementation Details}
We use pre-trained ViT-H SAM~\cite{kirillov2023segment} for weight initialization of the SAM. The resolution of input video frames is resized to $1024 \times 1024$.
We use AudioSet~\cite{gemmeke2017audio} pre-trained VGGish~\cite{hershey2017cnn} as an audio encoder. We set the frame length $T=5$, following~\cite{zhou2022audio}.
During training, the parameters of the SAM's image encoder and the audio backbone encoder are not updated.
We set the depth of ST-BAVA layers $L$ to 5 in S4 and 7 in MS3.
We set the attention order in ST-BAVA based on slightly better performance, but there was no significant difference.
The model is trained for 250 epochs, where Adam is used as an optimizer with a learning rate of $1e-4$.
Adapters that consist of two-layer MLP are inserted into all 32 image encoder layers of SAM, following the prior work~\cite{liu2023annotation}.
We train our model with Adapters for 15 epochs from separately pre-trained ST-BAVA and Adapters.

\section{RESULTS}
\label{sec:results}

\begin{figure*}[th!]
  \centering
     \begin{subfigure}[b]{0.087\textwidth}
        \includegraphics[trim={0cm 0cm 32.5cm 1cm}, clip=true, width=1.0\linewidth]{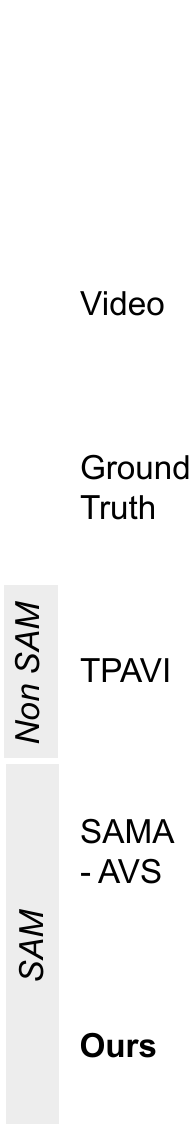}
        \vspace{1mm}
     \end{subfigure}
   \hfill
     \begin{subfigure}[b]{0.44\textwidth}
        \includegraphics[trim={3.8cm 0cm 15.9cm 1cm}, clip=true, width=1.0\linewidth]{figures/figure_qual_v6.pdf}
        \label{fig:S4}
        \vspace{-3mm}
        \caption{Single sound source segmentation (S4)}
        \vspace{-1mm}
     \end{subfigure}
   \hfill
     \begin{subfigure}[b]{0.433\textwidth}
        \includegraphics[trim={20cm 0cm 0cm 1cm}, clip=true, width=1.0\linewidth]{figures/figure_qual_v6.pdf}
        \label{fig:MS3}
        \vspace{-3mm}
        \caption{Multi sound sources segmentation (MS3)}
        \vspace{-1mm}
     \end{subfigure}
   \caption{Qualitative comparison with existing methods. Our method accurately identifies sound sources across multiple frames and describes detailed object shapes, achieving the most accurate segmentation performance.}
    \vspace{-1mm}
   \label{fig:results_qual}
\end{figure*}

\subsection{Quantitative Comparison}

Quantitative comparisons between our method and other state-of-the-art works on the AVS benchmark are presented in Table~\ref{tab:main_result_v2}.
Our proposed ST-BAVA with SAM outperforms the state-of-the-art methods on the AVSbench in all settings.
It achieves a significant performance gap in MS3, with 5.31 (8.3\%) mIoU improvement compared to the previous best score (GAVS~\cite{wang2023prompting}).
As MS3 presents the challenging task of distinguishing multiple objects corresponding to the sound within the image frame, achieving high performance on the MS3 indicates that the proposed model accurately identifies and localizes sound sources in the complex scene. 
We infer that our ST-BAVA supports SAM in leveraging the audio-visual correspondence between multiple sources aggregated in spatial and temporal dimensions.
Note that the trainable parameters in our model are small ($<$4\% of SAM), verifying its efficiency.

\subsection{Qualitative Examples}

Fig.~\ref{fig:results_qual} illustrates qualitative examples of the existing methods and ours.
Our model provides examples of accurately identifying the correct sound sources among visual candidate objects.
In the left video of the figure, our model successfully localizes the small sound source, the gun.
In contrast, SAMA-AVS~\cite{liu2023annotation}, another SAM-based approach without leveraging the temporal audio-visual relationship, incorrectly predicts the silent person as a sound source in several frames.
It supports our claim that the proposed approach benefits from utilizing temporal relationships to comprehend the scene with more contextual information, leading to the accurate distinction of the sounding objects among visual candidates.
Moreover, our model precisely delineates the gun's refined boundaries, including the bottom part of the gun.
Another non-SAM-based approach (TPAVI~\cite{zhou2022audio}) mispredicts the hand holding the gun or unrelated backgrounds.
In more challenging scenarios with multiple objects for the right video, our model correctly identifies the piano and the man as the sound sources and precisely portrays their shapes.

\subsection{Comparison to SAM with Visual Prompts}
\begin{table}[t]
  \resizebox{\linewidth}{!}{%
    \centering
    \small
    \renewcommand{\tabcolsep}{1mm}
    \begin{tabular}{ll|cccc}
    \hline
     \multirow{2}{*}{Approach} & \multirow{2}{*}{Methods} & \multicolumn{2}{c}{S4} & \multicolumn{2}{c}{MS3}  \\ \cline{3-6} 
      & & mIoU      & F-score    & mIoU      & F-score \\ \hline
    \multirow{2}{*}{Visual Prompts} & 1 Point & 42.45 & 0.637 & 33.01 & 0.523   \\
    \multirow{2}{*}{from G.T. mask} & 3 Points & 68.57 & 0.839 & 54.51 & 0.701   \\
    \multirow{2}{*}{without Training} & 1 Box & 76.01 & 0.867 & 63.46 & 0.713   \\
    & 3 Boxes & 76.75 & 0.874 & 66.76 & \textbf{0.813}  \\ \hline
    & w/o fusion module~\cite{liu2023annotation} & 81.53 & 0.886 & 63.14 & 0.691   \\
    & + TPAVI~\cite{zhou2022audio} & 81.68 & 0.902 & 64.78 & 0.749   \\
    Audio Prompts & + HAN~\cite{tian2020unified} & 80.56 & 0.896  & 64.14  & 0.739    \\
    with Training & + CMRAN~\cite{xu2020cross} & 81.46 & 0.899 & 65.09  & 0.747    \\
    & + JCA~\cite{praveen2022joint} & 81.99 & 0.903  & 65.44  & 0.751    \\
    & \cellcolor[HTML]{EFEFEF}+ ST-BAVA (Ours) & \cellcolor[HTML]{EFEFEF}\textbf{82.46} & \cellcolor[HTML]{EFEFEF}\textbf{0.906} & \cellcolor[HTML]{EFEFEF}\textbf{69.01} & \cellcolor[HTML]{EFEFEF}0.776 \\ \hline
    \end{tabular}
  }
  \caption{Comparison of different approaches to handle the AVS with SAM. 
  (Top) SAM receives point or box prompts guiding the region of the sound sources extracted from the ground-truth (G.T.) mask.
  (Bottom) For training SAM, intermediate feature fusion modules are inserted into SAM. 
  }
  \label{tab:abl_sam_appraoch}
  \vspace{-3mm}
\end{table}
% \begin{table}[t]
%   \resizebox{\linewidth}{!}{%
%     \centering
%     \small
%     \begin{tabular}{l|cccc}
%     \hline
%      \multirow{2}{*}{Methods} & \multicolumn{2}{c}{S4} & \multicolumn{2}{c}{MS3}  \\ \cline{2-5} 
%       & mIoU      & F-score    & mIoU      & F-score \\ \hline
%     \rowcolor[HTML]{EFEFEF}
%     SAM with GT (1 Point) & 42.45 & 0.637 & 33.01 & 0.523   \\
%     \rowcolor[HTML]{EFEFEF}
%     SAM with GT (3 Points) & 68.57 & 0.839 & 54.51 & 0.701   \\
%     \rowcolor[HTML]{EFEFEF}
%     SAM with GT (Box) & 76.01 & 0.867 & 63.46 & 0.713   \\
%     \rowcolor[HTML]{EFEFEF}
%     SAM with GT (Boxes) & 76.75 & 0.874 & 66.76 & 0.813  \\ \hline
%     SAM (w/o fusion module)  & 81.53 & 0.886 & 63.14 & 0.691   \\
%     SAM + CMRAN~\cite{xu2020cross} & - & - & -  & -    \\
%     SAM + HAN~\cite{tian2020unified} & -  & -  & -  & -    \\
%     SAM + TPAVI~\cite{zhou2022audio} & 81.68 & 0.902 & 64.78 & 0.749   \\
%     SAM + ST-BAVA (Ours) & \textbf{82.46} & \textbf{0.906} & \textbf{69.01} & \textbf{0.775} \\ \hline
%     \end{tabular}
%   }
%   \caption{Comparison with different approaches to handle the AVS with SAM. 
%   intermediate feature fusion modules inserted into SAM. 
%   % Adapter~\cite{liu2023annotation} is not used here for fair comparison.
%   }
%   \label{tab:abl_feature_fusion}
%   \vspace{-3mm}
% \end{table}
To verify the effectiveness of the proposed audio processing approach, we compare our model with the zero-shot performance of vanilla SAM using visual prompts on the AVS benchmark in Table~\ref{tab:abl_sam_appraoch}.
Since SAM does not inherently support audio prompts, we extract the points and boxes from the ground truth mask $M$ to guide the region of sound sources.
It is an alternative to user prompts employed in zero-shot approaches with SAM in various studies~\cite{mazurowski2023segment, cheng2023segment}.
In the case of points, we extract the largest regions corresponding to the sound sources and select the center of mass (or random if the center doesn't lie on the object) point per region.
For the boxes, the minimum external bounding boxes containing the largest contour region of sound sources in $M$ are used.

Results in Table~\ref{tab:abl_sam_appraoch} show that our model performs comparable to or even better than the zero-shot performance of SAM using the ground truth regions of sound sources as a visual prompt.
It verifies our method's effectiveness in handling AVS by adapting SAM with the direct capability to process audio without relying on manual prompts such as points and boxes.
Note that our approach leverages the temporal cross-modal context across the multi-frame, while the vanilla SAM with visual prompts does not. 

\subsection{Ablation on Feature Fusion Modules with SAM}
To prove the effectiveness of the ST-BAVA in adapting the SAM to AVS, we conducted an ablation study on the intermediate feature fusion module.
For comparison, we adopt the audio-visual feature fusion modules proposed in 
audio-visual segmentation (Temporal Pixel-wise Audio-Visual Interaction~\cite{zhou2022audio, mao2023multimodal, mao2023contrastive}),
audio-visual video parsing (Hybrid Attention Network~\cite{tian2020unified}),
audio-visual event localization (Cross-Modal Relation-Aware Networks~\cite{xu2020cross}),
and dimensional emotion recognition (Joint Cross-Attention~\cite{praveen2022joint}).

Results in Table~\ref{tab:abl_sam_appraoch} show that the proposed ST-BAVA outperforms other methods with SAM in the AVS benchmark.
% Without any intermediate module, SAM does not achieve sufficient performance as the 2-layer attention of SAM's decoder is too shallow to learn the complex audio-visual correspondences.
From the SAM without intermediate fusion module~\cite{liu2023annotation}, the performance improvement by introducing the intermediate fusion module demonstrates its significance in supporting the SAM's decoder to learn complex audio-visual correspondences.
However, all other fusion modules perform less than the ST-BAVA.
In the case of TPAVI~\cite{zhou2022audio}, the single integrated spatio-temporal attention is susceptible to implicit and redundant representations~\cite{bertasius2021space, huang2023discovering}, 
whereas the ST-BAVA separately operates spatial and temporal attention.
Furthermore, bidirectionally updating audio-visual features adopted by ST-BAVA enhances their subsequent cross-attention in the SAM's mask decoder.
Other modules, HAN~\cite{tian2020unified}, CMRAN~\cite{xu2020cross}, and JCA~\cite{praveen2022joint}, struggle with adapting SAM into the AVS, as they are not designed to consider the spatial visual features that are not essential to solving their tasks.

% \begin{table}[t]
%     \centering
%   \resizebox{\linewidth}{!}{%
%   \small
%     \begin{tabular}{cc|cccc}
%     \hline
%      &  &  \multicolumn{2}{c}{S4} & \multicolumn{2}{c}{MS3} \\ \cline{3-6} 
%     \multirow{-2}{*}{\begin{tabular}[c]{@{}c@{}}Temp.\end{tabular}} & \multirow{-2}{*}{\begin{tabular}[c]{@{}c@{}}Bi-dir.\end{tabular}} & mIoU & F-score & mIoU & F-score \\ \hline
%      - & - & 76.65 & 0.857 & 61.54 & 0.703  \\
%      \checkmark & - & 77.69 & 0.868 & 62.87 & 0.720 \\
%      - & \checkmark & 77.95 & 0.869 & 62.52 & 0.715 \\
%     \rowcolor[HTML]{EFEFEF} 
%      \checkmark & \checkmark & \textbf{80.02} & \textbf{0.888} & \textbf{66.06} & \textbf{0.743}  \\ \hline
%     \end{tabular}
%         }
%   \caption{Ablation on the temporal and bidirectional attention in the proposed ST-BAVA module. Temp. and Bi-dir. stand for Temporal and Bidirectional attention.}
%   \label{tab:result_abl_attn}
% \end{table}
\begin{table}[t]
    \centering
    \resizebox{\linewidth}{!}{
    % \small
    \begin{tabular}{l|cccc}
    \hline
    & \multicolumn{2}{c}{S4} & \multicolumn{2}{c}{MS3} \\  
    \multirow{-2}{*}{Methods} & mIoU & F-score & mIoU & F-score \\ \hline
    Baseline (Spatial A2V Attn.) & 76.65 & 0.857 & 61.54 & 0.703 \\ 
    w/o Temporal Attn. & 80.72 & 0.892 & 65.37 & 0.752 \\
    w/o Bidirectional Attn. & 80.09 & 0.887 & 65.17 & 0.749 \\
    w/o Adapter & 80.02 & 0.888 & 66.06 & 0.743 \\
    % w/o Adapter$^\dagger$ & 79.62 & 0.885 & 65.07 & 0.740 \\
    \rowcolor[HTML]{EFEFEF} 
    Full & \textbf{82.46} & \textbf{0.906} & \textbf{69.01} & \textbf{0.776} \\ \hline
    \end{tabular}
    }
    \caption{Ablation on the components of our methods.}
    \label{tab:result_abl_attn}
    \vspace{-3mm}
\end{table}

\subsection{Ablation on Model Components} 
\label{sec:res_abl_on_com}
We conduct an ablation study to investigate the effectiveness of the proposed components in Table~\ref{tab:result_abl_attn}.
The baseline uses spatial and unidirectional audio-to-image attention in ST-BAVA without Adapter~\cite{liu2023annotation}.
All of the proposed components yield performance improvements in both subsets in the AVS benchmark, highlighting the effectiveness of the ST-BAVA.
Notably, our proposed model without the Adapter performs well with no training or prompt-tuning of the SAM's image encoder. 
Utilizing the Adapter helps better cross-modal interaction in ST-BAVA with the audio-adapted image feature, further enhancing the AVS performance.

\begin{figure}[t]
  \centering
   \includegraphics[trim={8.3cm 2.5cm 8.7cm 4.7cm}, clip=true, width=0.97\linewidth]{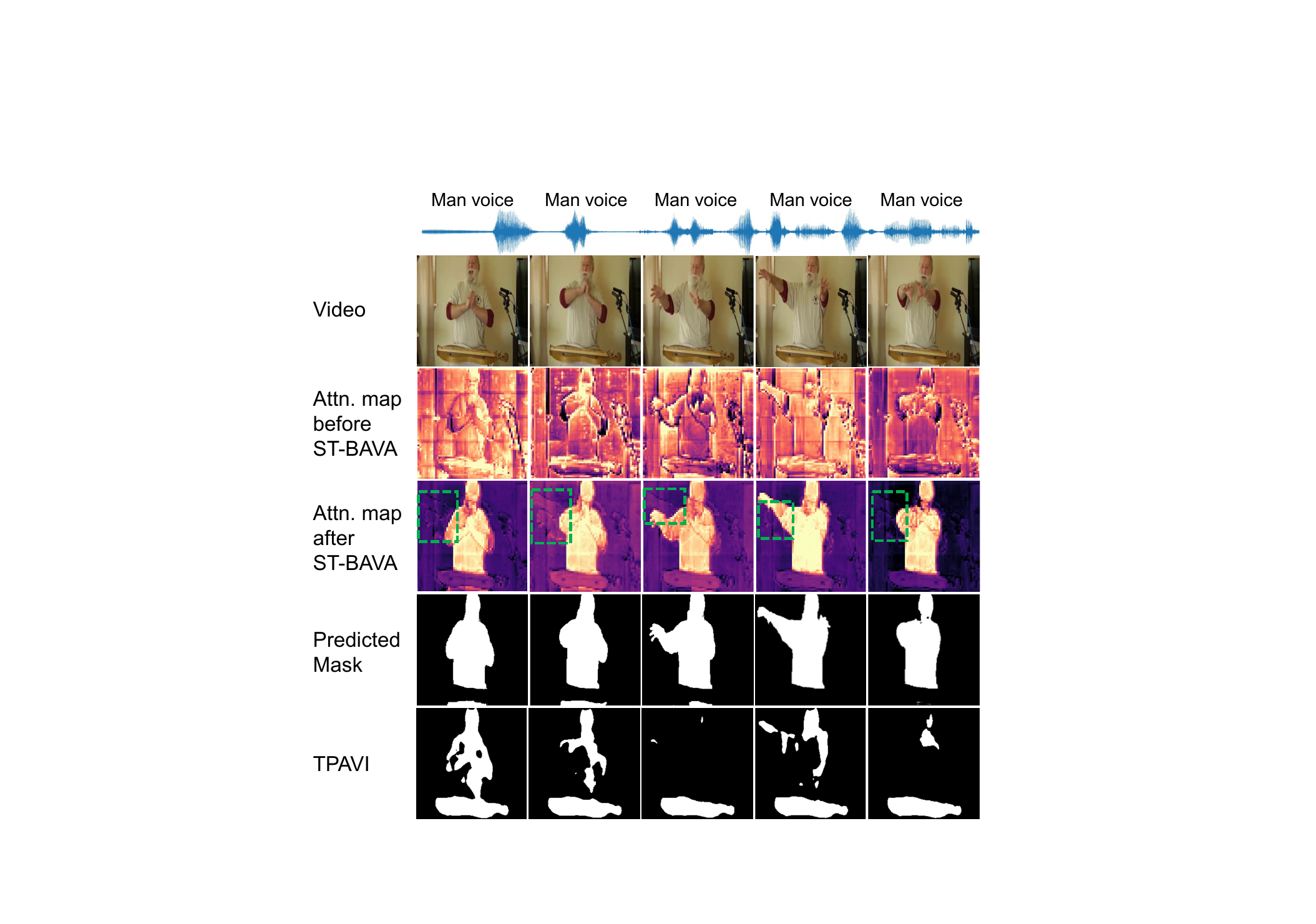}
   \caption{Spatial attention maps of the audio and visual embedding in the middle of our model pipeline.
   The attention map before ST-BAVA is calculated with the features extracted from the backbones.
   After the ST-BAVA, the map separately represents the region of sound sources within the frames, which leads to the correct segmentation of the sources in the predicted mask.
   Green-boxed regions show the visual information aggregated from other frames by temporal attention (the man with multiple arms).
   }
   \label{fig:results_feature_sim}
  \vspace{-3mm}
\end{figure}

\subsection{Analysis on Attention Maps}
To qualify the effect of the proposed ST-BAVA on the cross-modal features, we visualize the spatial attention score map between the audio-visual features in the middle of our model pipeline.
Since the mask decoder directly uses the attention map to get a final prediction mask, the intermediate maps present valuable cues affecting the model's output.

In Fig.~\ref{fig:results_feature_sim}, the attention maps before the ST-BAVA module do not include any information related to the sound sources, simply depicting the boundaries of objects on the image by the pre-knowledge of the backbones.
In contrast, the attention map after ST-BAVA clearly shows the high values in the sources' location, while the values in the backgrounds and the silent objects are low.
It leads to the accurate segmentation results observed in the predicted masks, showing the effectiveness of the ST-BAVA module in judging the pixel-level audio-visual correspondence.
Interestingly, after ST-BAVA, each map aggregates the visual information of other frames by temporal attention.
In the green-boxed region, the man on the map has multiple arms appearing in other frames, which does not disrupt the precise prediction masks at each time step.

\section{CONCLUSION}
\label{sec:conclusion}
We have proposed the auditory and temporal extension of SAM to solve the Audio-Visual Segmentation (AVS) task with audible video.
To comprehend the spatio-temporal relationship between the multiple image and audio frames, we have introduced the Spatio-Temporal, Bidirectional Audio-Visual Attention (ST-BAVA) module.
ST-BAVA is integrated between SAM's image encoder and mask decoder, aiming to deliver the audio-visual information exploited in spatial and temporal dimensions to SAM.
In the module, spatial attention calculates the pixel-wise audio-visual correspondence at each time step of the video, and temporal attention captures the contextual cross-modal relationship across the subsequent frames.
Extensive experimental results have verified that our model achieves meaningful performance in the AVS benchmark compared with state-of-the-art methods.

% \vfill\pagebreak
% \bibliographystyle{IEEEbib}
% \bibliography{main}

\end{document}